\documentclass[10pt,twocolumn,letterpaper]{article}

\usepackage{iccv}
\usepackage{times}
\usepackage{epsfig}
\usepackage{graphicx}
\usepackage{amsmath}
\usepackage{amssymb}

\usepackage{mathtools}
\makeatletter
\@namedef{ver@everyshi.sty}{}
\makeatother
\usepackage{tikz}
\usetikzlibrary{arrows}
\usepackage{verbatim}
\usepackage{xcolor,colortbl}
\usepackage{makecell}
\usepackage{multirow}

\usepackage[pagebackref=true,breaklinks=true,letterpaper=true,colorlinks,bookmarks=false]{hyperref}

\iccvfinalcopy % *** Uncomment this line for the final submission

\def\iccvPaperID{****} % *** Enter the ICCV Paper ID here

\ificcvfinal\fi

\begin{document}

%%%%%%%%% TITLE
\title{Data Augmentation for Scene Text Recognition}

\author{Rowel Atienza\\
Electrical and Electronics Engineering Institute\\
University of the Philippines\\
{\tt\small rowel@eee.upd.edu.ph}

}

\maketitle
\ificcvfinal\thispagestyle{empty}\fi

%%%%%%%%% ABSTRACT
\begin{abstract}
Scene text recognition (STR) is a challenging task in computer vision due to the large number of possible text appearances in natural scenes. Most STR models rely on synthetic datasets for training since there are no sufficiently big and publicly available labelled real datasets. Since STR models are evaluated using real data, the mismatch between training and testing data distributions results into poor performance of models especially on challenging text that are affected by noise, artifacts, geometry, structure, etc. In this paper, we introduce STRAug which is made of 36 image augmentation functions designed for STR. Each function mimics certain text image properties that can be found in natural scenes, caused by camera sensors, or induced by signal processing operations but poorly represented in the training dataset. When applied to strong baseline models using RandAugment, STRAug significantly increases the overall absolute accuracy of STR models across regular and irregular test datasets by as much as 2.10\% on Rosetta, 1.48\% on R2AM, 1.30\% on CRNN, 1.35\% on RARE, 1.06\% on TRBA and 0.89\% on GCRNN. The diversity and simplicity of API provided by STRAug functions enable easy replication and validation of existing data augmentation methods for STR. STRAug is available at \href{https://github.com/roatienza/straug}{https://github.com/roatienza/straug}.
\end{abstract}

\section{Introduction}

Humans use text to convey information through labels, signs, tags, logos, billboards and markers. For instance, a road sign with "Yield" informs drivers to wait for their turn. An "EXIT" sign points to the way going out. In the supermarket, sellers use labels or tags to inform buyers about the price and the quantity of products. Therefore, machines that read text in natural scenes can perform smarter decisions and actions.

The practical applications of scene text recognition (STR) have recently drawn interest in the computer vision community. Unfortunately, majority of the focus has always been on refining model architecture and training algorithm to improve the text recognition performance. While there is nothing wrong with this, STR can also benefit from the improvement in data for training. In the absence of sufficiently large and publicly available labelled datasets, the advancement of STR relies on huge collections of automatically annotated synthetic text images for training such as MJSynth or Synth90k \cite{jaderberg2014synthetic}, SynthText \cite{gupta2016synthetic}, Verisimilar \cite{zhan2018verisimilar}, and UnrealText \cite{long2020unrealtext}. Trained models are then evaluated on much smaller and fragmented real datasets such as IIIT5K (IIIT) \cite{mishra2012scene}, Street View Text (SVT) \cite{wang2011end}, ICDAR2003 (IC03) \cite{lucas2005icdar}, ICDAR2013 (IC13) \cite{karatzas2013icdar}, ICDAR2015 (IC15) \cite{karatzas2015icdar}, SVT Perspective (SVTP) \cite{phan2013recognizing} and CUTE80 (CT) \cite{risnumawan2014robust}. As a result, STR suffers from the typical problem of distribution shift from the training data to the evaluation data. STR models perform poorly especially in under represented or long tail samples similar to that can be found in the test data.

\begin{figure}
    \centering
\setlength\tabcolsep{1.5pt}
\begin{tabular}{ | c |  c |  c  | c | c | }
\hline
Model & \makecell{Input Text\\Image} & \makecell{Baseline \\ Prediction} & \makecell{+STRAug\\Prediction} &  \makecell{\%Acc \\ Gain}\\
\hline
CRNN \cite{shi2016end}& 
 \includegraphics[scale=0.23]{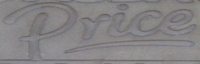} & P\textcolor{red}{y}ice & Price & 1.30 \\
\hline
R2AM \cite{lee2016recursive}& 
 \includegraphics[scale=0.23]{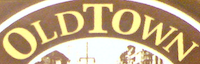} & OLDTOW & OLDTOWN & 1.48 \\
 \hline
GCRNN \cite{wang2017gated}& 
 \includegraphics[scale=0.24]{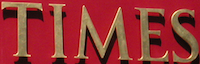} & TiM\textcolor{red}{m}ES & Times & 0.89 \\
\hline
Rosetta \cite{borisyuk2018rosetta}& 
 \includegraphics[scale=0.23]{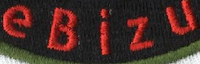} & eizu & eBizu & 2.10 \\
\hline
RARE \cite{shi2016robust} & 
 \includegraphics[scale=0.23]{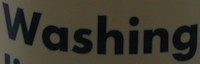} & Washi\textcolor{red}{i1} & Washing & 1.35 \\
\hline
TRBA \cite{baek2019wrong}& 
 \includegraphics[scale=0.23]{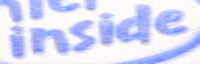} & insi\textcolor{red}{i}d & inside & 1.06 \\
\hline
\end{tabular}
    \caption{STRAug data augmentation significantly improves the overall accuracy of STR models especially on challenging input text images.  We follow the evaluation protocol used in most STR models of case sensitive training and case insensitive validation.}
    \label{fig:straug_teaser}
\end{figure}

In deep learning, popular approaches to address distribution shift include domain adaptation, causal representation learning, regularization, information bottleneck and data augmentation. In this paper, our focus is on the relatively straightforward approach of data augmentation. In STR, data augmentation has not been rigorously explored. Typical augmentation methods used include rotation, perspective and affine transformations, Gaussian noise, motion blur, resizing and padding, random or learned distortions, sharpening and cropping \cite{yu2020towards,lee2020recognizing,litman2020scatter,luo2020learn,aberdam2020sequence}. Proponents of STR methods select a subset of these augmentation methods to improve their models. To the best of our knowledge, there has been no comparative study on the performance of each of these methods. Furthermore, there are other possible text image augmentation functions that have not been used and fully explored in the existing STR literature.

In this paper, we attempt to formulate a library of data augmentation functions specifically designed for STR. While data augmentation algorithms are more developed in object recognition, they are not necessarily applicable in STR. For example, CutOut \cite{devries2017improved}, CutMix \cite{yun2019cutmix} and MixUp \cite{zhang2017mixup} can easily remove one or more symbols from the text image resulting in the total loss of useful information for training. In object recognition, there is generally only one class to predict. In STR, there are one or more characters each occupying a small region in the image. Removing a region or mixing two images will annihilate one or more characters in the text image. The correct meaning of the text could be altered.

STRAug proposes 36 augmentation functions designed for STR. Each function has a simple API:

\texttt{img = op(img, mag=mag, prob=prob)}. 

An STRAug function \texttt{op()} transforms an image \texttt{img}, with magnitude \texttt{mag} and with probability \texttt{prob}. Each function has 3 levels or magnitudes of severity or intensity that can manifest in capturing text in natural images. In order to avoid the combinatorial explosion in evaluating the effect of each augmentation function, we propose 8 logical groups based on the nature, origin or impact of these methods. The 8 groups are: 1) \textit{Warp}, 2) \textit{Geometry}, 3) \textit{Noise}, 4) \textit{Blur}, 5) \textit{Weather}, 6) \textit{Camera}, 7) \textit{Pattern} and 8) \textit{Process}. Using  RandAugment \cite{cubuk2020randaugment}, we demonstrate overall significant positive increase in text recognition accuracy of baseline models on both regular and irregular text datasets as shown in Figure \ref{fig:straug_teaser}. The simplicity of API and the number of functions supported by STRAug enable us to easily replicate and validate other data augmentation algorithms.

\begin{figure*}
    \centering
\setlength\tabcolsep{1.5pt}
\begin{tabular}{| c | c | c | c | c | c | c |}

\hline
Curved & Perspective & Shadow  & Noise & Pattern & Rotation & Stretched\\
 \includegraphics[scale=0.35]{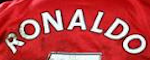} & 
 \includegraphics[scale=0.35]{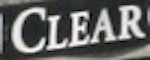} &
 \includegraphics[scale=0.35]{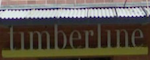} &
  \includegraphics[scale=0.35]{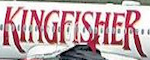} &
  \includegraphics[scale=0.35]{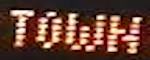} &
 \includegraphics[scale=0.35]{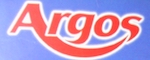} &
  \includegraphics[scale=0.35]{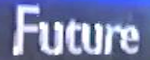}
  \\
  \hline
  
  \makecell{Uncommon \\ Font Style}  & \makecell{Blur and \\ Rotation}& \makecell{Occluded and \\ Curved} & \makecell{LowRes and \\ Pixelation} & \makecell{Distortion \\ and Rotation} & \makecell{Glass Reflect \\ \& Rotation} & \makecell{Uneven Light \\ \& Distortion} \\
 
 \includegraphics[scale=0.35]{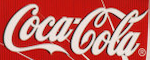} &
 \includegraphics[scale=0.35]{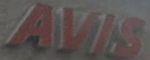} &
  \includegraphics[scale=0.35]{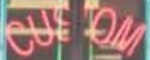} & \includegraphics[scale=0.35]{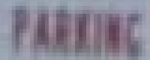}  &

   \includegraphics[scale=0.35]{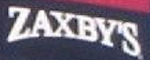} 
  & 
   \includegraphics[scale=0.35]{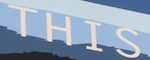} &
   \includegraphics[scale=0.35]{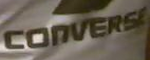}
  \\
  \hline
  
\end{tabular}

    \caption{Challenging text appearances encountered in natural scenes}
    \label{fig:text_variation}
\end{figure*}

\section{Related Work}

Scene text recognition (STR) is the challenging task of correctly reading a sequence of characters from natural images. STR models using deep learning \cite{yu2020towards,baek2019wrong,zhan2019esir,borisyuk2018rosetta,liu2016star} have superseded the performance of algorithms with hand-crafted features \cite{neumann2012real,yao2014unified}. Chen et al. \cite{chen2020text} presented a comprehensive review and analysis of different STR methods. The problem with deep learning models is that they require a large amount of data to automatically learn features. For STR, there are no publicly available large labelled real datasets. Collecting and annotating huge amount of real text data is a very costly and time consuming task. Thus, the advancement of STR relies on large synthetically generated and automatically annotated datasets.

Since STR models are evaluated on real, small and fragmented datasets, the bad side effects of data distribution shift are apparent especially on natural text images that are sometimes curved, noisy, distorted, blurry, under perspective transformation or rotated. In this paper, we believe that data augmentation can partially address the problem of data distribution shift in STR. Data augmentation automatically introduces certain transformations that can be found in the test datasets or natural scenes but under represented in the training datasets. Data augmentation can help in narrowing the gap between training and evaluation distributions. 

To the best of our knowledge, there has been no comprehensive study and empirical evaluation on different data augmentation functions that are helpful for STR models in general. Luo et al. \cite{luo2020learn} proposed \textit{Learn to Augment} to train an STR model to learn difficult text distortions. Experimental results on irregular text datasets such as ICDAR2015 (IC15) \cite{karatzas2015icdar}, SVT Perspective (SVTP) \cite{phan2013recognizing} and CUTE80 (CT) \cite{risnumawan2014robust}  demonstrated significant performance improvement. The disadvantage of \textit{Learn to Augment} is it requires additional agent and augmentation networks that must be trained with the main STR network. This results to a more complex setup, additional 1.5M network parameters, difficult to reuse algorithm and a longer training time. Furthermore, \textit{Learn to Augment} is only focused on distorted text, one of the many causes of data distribution shift in STR.

In the STR literature, data augmentation has been treated more of an after thought when proposing a new algorithm. Litman et al. \cite{litman2020scatter} used random resizing and distortion to improve SCATTER. Yu et al. \cite{yu2020towards} applied random resizing plus padding, rotation, perspective transformation, motion blur and Gaussian noise to gain additional performance for its semantic reasoning network (SRN). Lee et al. \cite{lee2020recognizing} used random rotation with a normal distribution to train its transformer-based SATRN model to improve irregular text recognition. Du et al. \cite{du2020pp} used a combination of techniques in \textit{Learn to Augment} and SRN to improve PP-OCR. While there is a strong evidence of improvement in performance, there is a lack of focused study in this area. In this paper, we attempt to address this problem by designing 36 data augmentation functions, creating 8 logical groups, analyzing the effect of each group, and systematically combining all groups to maximize their overall positive impact.

\section{Data Augmentation for STR}

Text in natural scenes can be found in various unconstrained settings such as walls, shirts, car plates, book covers, signboards, product labels, price tags, road signs, markers, etc. The captured text images have many degrees of variation in font style, orientation, shape, size, color, rendering, texture and illumination. The images are also subject to camera sensor orientation, location and imperfections causing image blur, pixelation, noise, and geometric and radial distortions. Weather disturbances such as glare, shadow, rain, snow and frost can also greatly affect the appearance of text. Figure \ref{fig:text_variation} shows that real-world text appearances are challenging for machines to read. In fact, text images may be simultaneously altered by several factors. In the following, we discuss the 36 functions classified into 8 groups that attempt to mimic the issues in capturing text in natural scenes.

\tikzstyle{image}=[minimum size=4em]

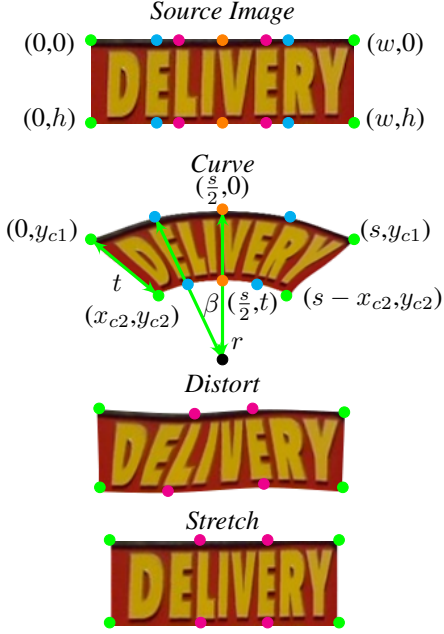
\begin{figure}[t]
\begin{center}

\begin{tikzpicture}[node distance=2cm,auto,>=latex']
    \node [image](source)[node distance=0.5em]
    {\includegraphics[width=.2\textwidth]{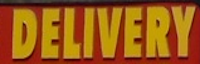}};

    \node [] (origin0) [left of=source, node distance=2.3cm, yshift=0.5cm]{(0,0)};
    
    \node [] (origin1) [right of=source, node distance=2.3cm, yshift=0.5cm]{($w$,0)};

    \node [] (origin2) [left of=source, node distance=2.3cm, yshift=-0.5cm]{(0,$h$)};
    
    \node [] (origin3) [right of=source, node distance=2.3cm, yshift=-0.5cm]{($w$,$h$)};
    
    \filldraw [green] (1.75,0.55) circle (2pt);
    \filldraw [green] (-1.75,0.55) circle (2pt);
    \filldraw [green] (1.75,-0.55) circle (2pt);
    \filldraw [green] (-1.75,-0.55) circle (2pt);
    \filldraw [orange] (0,0.55) circle (2pt);
    \filldraw [orange] (0,-0.55) circle (2pt);
    
    \filldraw [cyan] (.875,0.55) circle (2pt);
    \filldraw [cyan] (-.875,0.55) circle (2pt);

    \filldraw [cyan] (.875,-0.55) circle (2pt);
    \filldraw [cyan] (-.875,-0.55) circle (2pt);

    \filldraw [magenta] (.58,-0.55) circle (2pt);
    \filldraw [magenta] (-.58,-0.55) circle (2pt);
    \filldraw [magenta] (.58,0.55) circle (2pt);
    \filldraw [magenta] (-.58,0.55) circle (2pt);
    
    \node [image](curve)[below of=source, node distance=6.5em]
    {\includegraphics[width=.2\textwidth]{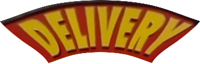}};
    
    \node [] (c1) [left of=curve, node distance=2.4cm, yshift=0.3cm]{(0,$y_{c1}$)};

    \node [] (c2) [left of=curve, node distance=1.2cm, yshift=-0.85cm]{($x_{c2}$,$y_{c2}$)};
    
    \node [] (center) [below of=source, node distance=4.4cm] {};

    \draw[<->, color=green, very thick] (0,-3.7) -- (0, -1.7);
    
    \draw[<->, color=green, very thick] (0,-3.7) -- (-.9, -1.8);
    
    \draw[<->, color=green, very thick] (-.85,-2.85) -- (-1.75, -2.1);
    
    \node [] (c3) [right of=curve, node distance=2.3cm, yshift=0.3cm]{($s$,$y_{c1}$)};

    \node [] (c4) [right of=curve, node distance=2cm, yshift=-0.6cm]{($s-x_{c2}$,$y_{c2}$)};
    
    %\draw[<->, color=white, very thick] (0.1,-2.85) -- (0.1, -1.7);
    
    \node [] (t) [left of=curve, node distance=1.4cm, yshift=-0.4cm]{$t$};
    
    \node [] (r) [below of=curve, node distance=1.2cm,xshift=0.2cm]{$r$};

    \node [] (curve_name) [above of=curve, node distance=2.4em]{($\frac{s}{2}$,0)};
    
    \node [] (mid) [below of=curve, node distance=0.7cm, xshift=.4cm]{ ($\frac{s}{2}$,$t$) };
    
    % high
    \filldraw [green] (-1.75, -2.1) circle (2pt);
    \filldraw [green] (1.75, -2.1) circle (2pt);
    
    % low
    \filldraw [green] (-.85,-2.85) circle (2pt);
    \filldraw [green] (.85,-2.85) circle (2pt);
    
    % mid
    \filldraw [cyan] (-.9, -1.8) circle (2pt); \filldraw [cyan] (.9, -1.8) circle (2pt);  
    % mid
    \filldraw [cyan] (-.46, -2.7) circle (2pt); \filldraw [cyan] (.46, -2.7) circle (2pt);

    %\draw[thick, <->] (0,-1.6cm) arc (0:10:10);
    
        % bottom and center
    \filldraw [black] (0,-3.7) circle (2pt);
    \filldraw [orange] (0,-1.7) circle (2pt);
    \filldraw [orange] (0,-2.65) circle (2pt);

    \node [] (beta) [left of=r, node distance=0.35cm, yshift=0.5cm]{$\beta$};
    
    \node [] (curve_name) [above of=curve, node distance=1.2cm]{\textit{Curve}};
    
    %\draw[->, color=orange, very thick] (0,-4.2) -- (1.75, -2);
    
    \node [image](distort)[below of=curve, node distance=7.5em]
    {\includegraphics[width=.2\textwidth]{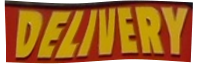}};
    
    %\node [] (d1) [left of=distort, node distance=2.5cm, yshift=0.5cm]{($x_{d1},y_{d1}$)};

    %\node [] (d1) [right of=distort, node distance=2.5cm, yshift=0.5cm]{($x_{d4},y_{d4}$)};
    
    %\node [] (d1) [left of=distort, node distance=2.5cm, yshift=-0.5cm]{($x_{d5},y_{d5}$)};    
    
    %\node [] (d1) [right of=distort, node distance=2.5cm, yshift=-0.5cm]{($x_{d8},y_{d8}$)};    
    
    % distort pts
    \filldraw [green] (1.62,-4.4) circle (2pt);
    \filldraw [green] (1.6,-5.4) circle (2pt);
    \filldraw [green] (-1.64,-4.35) circle (2pt);
    \filldraw [green] (-1.63,-5.4) circle (2pt);
    \filldraw [magenta] (.4,-4.35) circle (2pt);
    \filldraw [magenta] (.55,-5.355) circle (2pt);    
    \filldraw [magenta] (-.37,-4.42) circle (2pt);
    \filldraw [magenta] (-.725,-5.45) circle (2pt);        
    \node [image] (stretch)[below of=distort, node distance=5em]
    {\includegraphics[width=.2\textwidth]{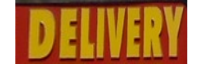}};

    %strecth
    \filldraw [green] (1.55,-6.1) circle (2pt);
    \filldraw [green] (1.55,-7.2) circle (2pt);
    \filldraw [green] (-1.5,-6.1) circle (2pt);
    \filldraw [green] (-1.5,-7.2) circle (2pt);
    
    \filldraw [magenta] (.6,-7.2) circle (2pt); \filldraw [magenta] (-.29,-7.2) circle (2pt);
    \filldraw [magenta] (.6,-6.1) circle (2pt);
    \filldraw [magenta] (-.3,-6.1) circle (2pt);

    %\node [] (d1) [left of=stretch, node distance=2.5cm, yshift=0.5cm]{($x_{s1},y_{s1}$)};

    %\node [] (d1) [right of=stretch, node distance=2.5cm, yshift=0.5cm]{($x_{s4},y_{s4}$)};
    
    %\node [] (d1) [left of=stretch, node distance=2.5cm, yshift=-0.5cm]{($x_{s5},y_{s5}$)};    
    
    %\node [] (d1) [right of=stretch, node distance=2.5cm, yshift=-0.5cm]{($x_{s8},y_{s8}$)};   

    \node [] (source_name) [above of=source, node distance=2.6em]{\textit{Source Image}};

    \node [] (distort_name) [above of=distort, node distance=.9cm]{\textit{Distort}};
    
    \node [] (distort_name) [above of=stretch, node distance=2.4em]{\textit{Stretch}};
    %\draw[->] (image3.east) -- (encoder3.west);
    %\draw[->] (encoder3.east) -- (output3.west);
    %\node [] (label3) [above of=encoder3, node distance=1.9em]{Transformer Encoder (ViTSTR)};  

\end{tikzpicture}

\end{center}
   \caption{Source and destination control points used in TPS image warping transformation for \textit{Curve}, \textit{Distort} and \textit{Stretch} data augmentation.}
\label{fig:warp}
\end{figure}

\begin{table}[]
    \centering
    \begin{tabular}{|c|c|}
    \hline
        Source Pt & Destination Pt \\
    \hline
    \multicolumn{2}{|c|}{\textit{Curve}} \\
    \hline
        $(0,0)$ &  \makecell{$(0,y_{c1})$ s.t. \\ $r=rand(r_{min},r_{max}){h}$ \\ $x_{1}=(r^2 -\frac{s^2}{4})^\frac{1}{2}$,   $y_{c1}=r- x_{1}$} \\ 
    \hline
        $(\frac{s}{4},0)$ & \makecell{$(x_{2},y_{2})=(\frac{s}{2}-r\sin{\beta},r(1-\cos{\beta}))$ s.t.\\
        $\sin{\beta}=(\frac{1}{2}-\frac{x_{1}}{2r})^\frac{1}{2}$,        $\cos{\beta}=(\frac{1}{2}+\frac{x_{1}}{2r})^\frac{1}{2}$
        }
        \\ 
    \hline
        $(\frac{3s}{4},0)$ & $(s-x_{2},y_{2})$
        \\
    \hline
        $(s,0)$ & $(s,y_{c1})$ \\ 
        
    \hline
        $(\frac{s}{4},s)$ & 
        \makecell{ $(x_{3},y_{3})=$ \\
        $(\frac{w}{2}-(r-t)\sin{\beta},r-(r-t)\cos{\beta})$} \\ 
    \hline
        $(\frac{3s}{4},s)$ & $(s-x_{3},y_{3})$ \\     
    \hline
        $(\frac{s}{2},0)$ & $(\frac{s}{2},0)$\\ 
    \hline
        $(\frac{s}{2},s)$ & $(\frac{s}{2},t)$ s.t. $t=\frac{s}{2}rand(0.4, 0.5)$\\ 
    \hline
        $(0,s)$ & \makecell{ $(x_{c2},y_{c2})$ s.t. \\ $x_{c2}=\frac{st}{2r}$, $y_{c2}=y_{c1}+\frac{tx_{1}}{r}$ }\\ 
    \hline
        $(s,s)$ & $(s-x_{c2},y_{c2})$ \\ 
    \hline
        \multicolumn{2}{|c|}{\textit{Distort} and \textit{Stretch}} \\
    \hline
        $(0,0)$ & $(\frac{w}{3}rand(0,k),\frac{h}{2}rand(0,k))$  \\ 
    \hline
        $(\frac{w}{3},0)$ &  $(\frac{w}{3}(1+rand(-k,k)),\frac{h}{2}rand(0,k))$  \\ 
    \hline
        $(\frac{2w}{3},0)$ & $(\frac{w}{3}(2+rand(-k,k)),\frac{h}{2}rand(0,k))$  \\ 
    \hline
        $(w,0)$ & $(w-\frac{w}{3}rand(0,k),\frac{h}{2}rand(0,k))$  \\ 
    \hline
        $(0,h)$ & $(\frac{w}{3}rand(0,k), h-\frac{h}{2}rand(0,k))$  \\ 
    \hline
        $(\frac{w}{3},h)$ & $(\frac{w}{3}(1+rand(-k,k)),h-\frac{h}{2}rand(0,k))$  \\ 
    \hline
        $(\frac{2w}{3},h)$ &  $(\frac{w}{3}(2+rand(-k,k)), h-\frac{h}{2}rand(0,k))$  \\ 
    \hline
        $(w,h)$ & $(w-\frac{w}{3}rand(0,k)), h-\frac{h}{2}rand(0,k))$  \\ 
    \hline
    \end{tabular}
    \caption{Formula for source and destination control points used by TPS image warping. For \textit{Curve}, the image is first resized to a square with side $s$ before the TPS transformation is applied. Afterward, the image is returned to its original dimensions $(w,h)$. $r$ decreases with the level of severity. \textit{Distort} and \textit{Stretch} share the same set of formula. For \textit{Stretch}, the $y$ coordinate is not randomized. $k$ increases with the level of intensity.}
    \label{tab:warp_formula}
\end{table}

\subsection{Warp}
Curved, distorted and stretched text styles are found in natural scenes but are usually not well represented in train datasets. The \textit{Warp} group includes \textit{Curve}, \textit{Distort} and \textit{Stretch}. Curved text images are found in logos, seals, coins, product labels, emblems and tires. Distorted text can be seen on clothing, textiles, candy wrappers, thin plastic packaging and flags. Stretched text can be observed on elastic packaging materials and balloons. In this paper, we use stretch to refer to elastic deformation which may include its literal opposite word meaning, contract. Both \textit{Distort} and \textit{Stretch} are also used in certain artistic styles or can be caused by structural deformation and camera lens radial distortion (e.g. fish eye lens).

Figure \ref{fig:warp} shows the control points used in smooth deformation of a horizontal text image into \textit{Curve}, \textit{Distort} and \textit{Stretch} versions.  We use Thin-Plate-Spline (TPS) \cite{bookstein1989principal}  to produce a warped version of the original image by moving pixels at source control points to their destination coordinates. All neighboring pixels around source control points are also re-positioned while following the smooth deformation constraints. With proper values of destination control points, various realistic deformations can be approximated such as curved, distorted and stretched. Table \ref{tab:warp_formula} lists the source and destination control points for our \textit{Warp} data augmentation. An alternative algorithm to TPS is moving least squares as used in STR by Luo et al. \cite{luo2020learn}. 

In the \textit{Curve} image warping, the text image is first resized to a square with side $s$. Then, random vertical flip is applied to get either concave or convex text shape. After the TPS smooth deformation, the upper half of the image is cropped since the lower half is covered by blank filler color. Then, the original image dimension is restored. We used 8 control points for warping. Two optional control points at the mid point of each side of the source image help improve the straightness of the edges. As the magnitude of augmentation increases, the radius of curvature $r$ decreases. For \textit{Distort} and \textit{Stretch}, the extent of distortion $k$ increases with the level of severity.

\begin{figure}[t]
\begin{center}

\begin{tikzpicture}[node distance=2cm,auto,>=latex']
    \node [image](source)[]{\includegraphics[width=.2\textwidth]{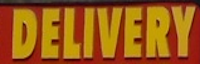}};

    \node [image](perspective)[right of=source, node distance=4cm]{\includegraphics[width=.2\textwidth]{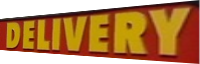}};    

    \node [image](shrink)[below of=source, node distance=2cm]{\includegraphics[width=.2\textwidth]{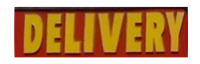}};    

    \node [image](rotate)[right of=shrink, node distance=4cm]{\includegraphics[width=.2\textwidth]{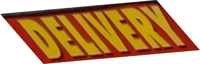}};    
%source
    \filldraw [green] (1.75,0.55) circle (2pt);
    \filldraw [green] (-1.75,0.55) circle (2pt);
    \filldraw [green] (1.75,-0.55) circle (2pt);
    \filldraw [green] (-1.75,-0.55) circle (2pt);

%perspective
    \filldraw [green] (5.75,0.55) circle (2pt);
    \filldraw [green] (2.25,0.25) circle (2pt);
    \filldraw [green] (5.75,-0.55) circle (2pt);
    \filldraw [green] (2.25,-0.35) circle (2pt);

%shrink
    \filldraw [green] (1.55,-1.55) circle (2pt);
    \filldraw [green] (-1.5,-1.55) circle (2pt);
    \filldraw [green] (1.55,-2.45) circle (2pt);
    \filldraw [green] (-1.5,-2.45) circle (2pt);
    
% rotate
\draw[ultra thick, ->, green] (4.5,-2) arc (0:330:.4);

    \node [] (source_name) [above of=source, node distance=.8cm]{\textit{Source Image}};
    
    \node [] (perspective_name) [above of=perspective, node distance=.8cm]{\textit{Perspective}};
    
    \node [] (shrink_name) [above of=shrink, node distance=.8cm]{\textit{Shrink}};
    
    \node [] (rotate_name) [above of=rotate, node distance=.8cm]{\textit{Rotate}};
    
    \node [] (theta) [below of=rotate, node distance=.8cm]{$\theta$};

\end{tikzpicture}

\end{center}
   \caption{Image transformation under \textit{Geometry} group.}
\label{fig:geometry}
\end{figure}
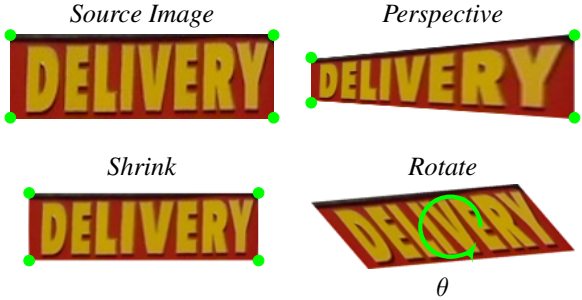

\begin{table}[]
    \centering
    \begin{tabular}{|c|c|}
    \hline
        Source Pt & Destination Pt \\
    \hline
        \multicolumn{2}{|c|}{\textit{Perspective}} \\
    \hline
        $(0,0)$ & $(0,h\cdot{rand(k,k+.1)})$  \\ 
    \hline
        $(w,0)$ & $(w,0)$  \\ 
    \hline
        $(0,h)$ & $(0,h\cdot{rand(0.9-k,1-k)})$  \\ 
    \hline
        $(w,h)$ & $(w,h)$ \\ 
    \hline
    \multicolumn{2}{|c|}{\textit{Shrink}} \\
    \hline
        $(0,0)$ & \makecell{$(\Delta{w},\Delta{h})$
        \\$\Delta{w}=\frac{w}{3}{rand(k,k+.1)}$ \\$\Delta{h}=\frac{h}{2}{rand(k,k+.1)}$}  \\ 
    \hline
        $(w,0)$ & $(w-\Delta{w},\Delta{h})$  \\ 
    \hline
        $(0,h)$ & $(\Delta{w},h-\Delta{h})$  \\ 
    \hline
        $(w,h)$ & $(w-\Delta{w},h-\Delta{h})$ \\ 
    \hline    
    \multicolumn{2}{|c|}{\textit{Rotate}, $\theta=rand(\theta_{min},\theta_{max})$} \\
    \hline    
    \end{tabular}
    \caption{Formula for source and destination control points used in \textit{Perspective} and \textit{Shrink} data augmentations. $k$ increases with the level of intensity. For \textit{Rotate}, $\theta$ is sampled from a uniform distribution. $\theta_{min}$ and $\theta_{max}$ values increase with the magnitude of data augmentation. }
    \label{tab:geometry_formula}
\end{table}

\subsection{Geometry}
When viewing natural scenes, perfect horizontal frontal alignment is seldom achieved. Almost always there is some degree of rotation and perspective transformation in the text image. Text may not also be perfectly centered. Translation along $x$ and/or $y$ coordinates is common. Furthermore, text can be found in varying sizes. To simulate these real-world scenarios, the \textit{Geometry} group includes \textit{Perspective}, \textit{Shrink} and \textit{Rotate} image transformations. Figure \ref{fig:geometry} shows these data augmentations while Table \ref{tab:geometry_formula} lists the source and destination control points. For \textit{Rotate}, there is only $\theta$ as the degree of freedom.

For \textit{Perspective}, the horizon can be at the left or right side of the image. For simplicity, Figure \ref{fig:geometry} and Table \ref{tab:geometry_formula} show left horizon only. For both \textit{Perspective} and \textit{Shrink}, $k$ increases with the magnitude of augmentation. We also use TPS to perform \textit{Shrink} deformation. As mentioned earlier, text may not be necessarily centered at all times. Therefore for \textit{Shrink}, we randomly translate along horizontal or vertical axis. To avoid unintentional cropping of text symbols, the maximum horizontal translation is set to $\Delta{w}$ (maximum of $\Delta{h}$ for vertical translation).

For \textit{Rotation}, $\theta$ is uniformly sampled from $\theta_{min}$ to $\theta_{max}$. The magnitude of data augmentation increases with $\theta_{min}$ and $\theta_{max}$. Clockwise rotation is supported by flipping the sign of $\theta$ with 50\% probability.

\begin{figure}
    \centering
\setlength\tabcolsep{1.5pt}
\begin{tabular}{ c  c  c }
\textit{Source Image} & \textit{Grid} & \textit{VGrid} \\
 \includegraphics[scale=0.35]{LaTeX/images/source/delivery.png} & 
 \includegraphics[scale=0.35]{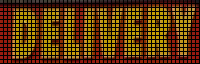} & 
 \includegraphics[scale=0.35]{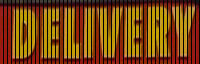} 
 \\
\textit{HGrid} & \textit{RectGrid} & \textit{EllipseGrid} \\

 \includegraphics[scale=0.35]{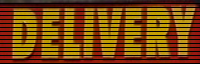} &
 \includegraphics[scale=0.35]{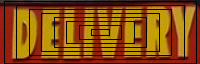} & 
 \includegraphics[scale=0.35]{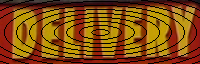}  
 \\
\end{tabular}
    \caption{Examples of images affected by \textit{Pattern} data augmentation.}
    \label{fig:pattern}
\end{figure}

\subsection{Pattern}
Regional dropout data augmentation methods such as CutOut \cite{devries2017improved}, MixUp \cite{zhang2017mixup} and CutMix \cite{yun2019cutmix} are not suitable in STR since one or more symbols may be totally removed from the image. Inspired by GridMask \cite{chen2020gridmask}, we designed 5 grid patterns that mask out certain regions from the image while ensuring that text symbols are still readable. For the \textit{Pattern} group, we introduce 5 types of Grid: \textit{Grid}, \textit{VGrid}, \textit{HGrid}, \textit{RectGrid} and \textit{EllipseGrid} as shown in Figure \ref{fig:pattern}. The distance between grid lines decreases with the magnitude of data augmentation. Text with grid like appearance can be found in certain electronic displays and billboards, dot-matrix printer type of fonts and signs behind a meshed fence.

\begin{figure}
    \centering
\setlength\tabcolsep{1.5pt}
\begin{tabular}{ c c  c }
\textit{Image Source} & \textit{GaussianNoise} & \textit{ShotNoise} \\
 \includegraphics[scale=0.35]{LaTeX/images/source/delivery.png} &
 \includegraphics[scale=0.35]{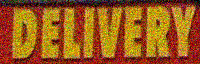} & 
 \includegraphics[scale=0.35]{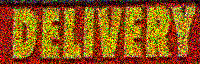} \\
\textit{ImpulseNoise} & \textit{SpeckleNoise} & {} \\
 \includegraphics[scale=0.35]{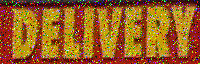} & 
 \includegraphics[scale=0.35]{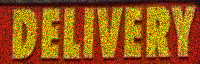} & {} \\ 
\end{tabular}
    \caption{Examples of images affected by \textit{Noise} data augmentation.}
    \label{fig:noise}
\end{figure}

\subsection{Noise}
Noise is common in natural images. For STRAug, we lumped together different \textit{Noise} types: 1) \textit{GaussianNoise}, 2) \textit{ShotNoise}, 3) \textit{ImpulseNoise} and 4) \textit{SpeckleNoise}. Figure \ref{fig:noise} shows how each type of noise affects the text image. Gaussian noise manifests in low-lighting conditions. Shot noise or Poisson noise is electronic noise due to the discrete nature of light itself. Impulse noise is a color version of salt-and-pepper noise which can be caused by bit errors. For the \textit{Noise} group, we adopted the implementation by Hendrycks and Dietterich \cite{hendrycks2018benchmarking} but using only half of the levels in order to ensure that the text in the image is still human readable. The amount of noise corruption increases with the level of severity of data augmentation.

\begin{figure}
    \centering
\setlength\tabcolsep{1.5pt}
\begin{tabular}{ c  c  c }
\textit{Source Image} & \textit{GaussianBlur} & \textit{DefocusBlur}\\
 \includegraphics[scale=0.35]{LaTeX/images/source/delivery.png} & 
 \includegraphics[scale=0.7]{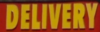} &
 \includegraphics[scale=0.7]{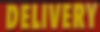} \\
\textit{MotionBlur} & \textit{GlassBlur} & \textit{ZoomBlur}\\
 \includegraphics[scale=0.7]{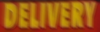} & 
 \includegraphics[scale=0.7]{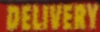} &
  \includegraphics[scale=0.7]{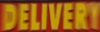} 
 \\
\end{tabular}
    \caption{Example images affected by \textit{Blur} data augmentation.}
    \label{fig:blur}
\end{figure}

\subsection{Blur}
Similar to noise, blur is common in natural images. Blur may be caused by unstable camera sensor, dirty lens, relative motion between the camera and the subject, insufficient illumination, out of focus settings, imaging while zooming, subject behind a frosted glass window, or shallow depth of field. The \textit{Blur} group includes: 1) \textit{GaussianBlur}, 2) \textit{DefocusBlur}, 3) \textit{MotionBlur}, 4) \textit{GlassBlur} and 5) \textit{ZoomBlur}. Figure \ref{fig:blur} shows resulting images due to \textit{Blur} functions. The degree of blurring increases with the level of severity of data augmentation. Except for \textit{GaussianBlur}, we adopted the implementation by Hendrycks and Dietterich \cite{hendrycks2018benchmarking}.

\begin{figure}
    \centering
\setlength\tabcolsep{1.5pt}
\begin{tabular}{ c  c  c }
\textit{Source Image} & \textit{Fog} & \textit{Snow}\\
 \includegraphics[scale=0.35]{LaTeX/images/source/delivery.png} & 
 \includegraphics[scale=0.7]{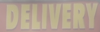} &
 \includegraphics[scale=0.7]{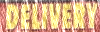} \\
\textit{Frost} & \textit{Rain} & \textit{Shadow}\\
 \includegraphics[scale=0.7]{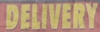} & 
 \includegraphics[scale=0.7]{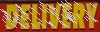} &
  \includegraphics[scale=0.7]{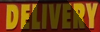} 
 \\
\end{tabular}
    \caption{Example images affected by \textit{Weather} data augmentation.}
    \label{fig:weather}
\end{figure}

\subsection{Weather}
Scene text may be captured under different weather conditions. As such, we simulate these conditions under \textit{Weather} group: 1) \textit{Fog}, 2) \textit{Snow}, 3) \textit{Frost}, 4) \textit{Rain} and 5) \textit{Shadow}. Figure \ref{fig:weather} shows how a text image is affected by different weather conditions. As the magnitude of data augmentation increases, the severity of weather condition is increased. For example, as the magnitude of augmentation increases, the number of rain drops increases or the opacity of the shadow increases. The weather conditions around the world are extremely varied that it may not be possible to cover all possible scenarios. \textit{Weather} simulates some common conditions only.

\begin{figure}
    \centering
\setlength\tabcolsep{1.5pt}
\begin{tabular}{ c  c  c }
\textit{Source Image} & \textit{Contrast} & \textit{Brightness}\\
 \includegraphics[scale=0.35]{LaTeX/images/source/delivery.png} & 
 \includegraphics[scale=0.7]{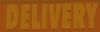} &
 \includegraphics[scale=0.7]{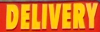} \\
\textit{JpegCompression} & \textit{Pixelate} & {}\\
 \includegraphics[scale=0.7]{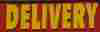} & 
 \includegraphics[scale=0.7]{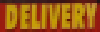} &
  {} 
 \\
\end{tabular}
    \caption{Example images affected by \textit{Camera} data augmentation.}
    \label{fig:camera}
\end{figure}

\subsection{Camera}
Camera sensors have many imperfections and tunable settings. These are grouped under \textit{Camera}: 1) \textit{Contrast}, 2) \textit{Brightness}, 3) \textit{JpegCompression} and 4) \textit{Pixelate}. \textit{Contrast} enables us to distinguish the different objects that compose an image. This could be the text against background and other artifacts. \textit{Brightness} is directly affected by scene luminance. \textit{JpegCompression} is the side effect of image compression. \textit{Pixelate} is exhibited by increasing the resolution of an image. The severity of camera effect increases with the level of data augmentation. Figure \ref{fig:camera} illustrates the effect of \textit{Camera} data augmentation.

\begin{figure}
    \centering
\setlength\tabcolsep{1.5pt}
\begin{tabular}{ c  c  c }
\textit{Source Image} & \textit{Posterize} & \textit{Solarize}\\
 \includegraphics[scale=0.35]{LaTeX/images/source/delivery.png} & 
 \includegraphics[scale=0.7]{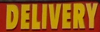} &
 \includegraphics[scale=0.7]{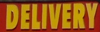} \\
\textit{Invert} & \textit{Equalize} & \textit{AutoContrast}\\
 \includegraphics[scale=0.7]{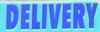} & 
 \includegraphics[scale=0.7]{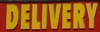} &
  \includegraphics[scale=0.7]{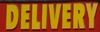} 
 \\
 \textit{Sharpness} & \textit{Color} & \textit{}\\
 \includegraphics[scale=0.7]{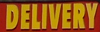} & 
 \includegraphics[scale=0.7]{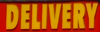} &
  {} 
 \\
\end{tabular}
    \caption{Example images affected by \textit{Process} data augmentation.}
    \label{fig:process}
\end{figure}

\subsection{Process}
All other image transformations used in object recognition data augmentation literature that may be applicable in STR are grouped together in \textit{Process}: 1) \textit{Posterize}, 2) \textit{Solarize}, 3) \textit{Invert}, 4) \textit{Equalize}, 5) \textit{AutoContrast}, 6) \textit{Sharpness} and 7) \textit{Color}. Figure \ref{fig:process} demonstrates the effect of \textit{Process}. These functions were used in AutoAugment \cite{cubuk2019autoaugment} and can help STR models learn invariant features of text in images. The functions are image processing routines that change the image appearance but not the readability of the text. This is done through bit-wise or color manipulation. For example, \textit{Invert} and \textit{Color} can drastically change the color of the image but the readability of the text remains.  \textit{Invert}, \textit{AutoContrast} and \textit{Equalize} support 1 level of intensity only. 

\section{Experimental Results and Discussion}

We evaluated the impact of STRAug on different strong baseline STR methods using the framework developed by Baek et al. \cite{baek2019wrong}. We first describe the train and test datasets. Then, we present and analyze the empirical results.

\subsection{Train Dataset}

\begin{figure}
    \centering
\setlength\tabcolsep{1pt}
\begin{tabular}{| c | c  |}
\hline
MJ & ST \\
\hline
\includegraphics[scale=0.3]{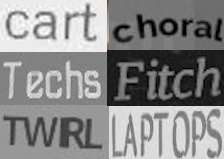} & \includegraphics[scale=0.3]{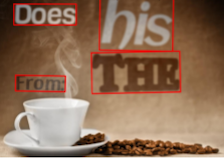} \\
\hline

\end{tabular}    
    \caption{Sample images from train datasets.}
    \label{fig:synth_dataset}
\end{figure}

The framework uses 1) MJSynth (MJ) \cite{jaderberg2014synthetic} or also known as Synth90k and 2) SynthText (ST) \cite{gupta2016synthetic} to train STR models. Figure \ref{fig:synth_dataset} shows sample images from MJ and ST. We provide a short description of the two datasets.

\textbf{MJSynth (MJ)} is a synthetically generated dataset made of 8.9M realistically looking  word images. MJSynth was designed to have 3 layers: 1) background, 2) foreground and 3) optional shadow/border. It uses 1,400 different fonts, different background effects, border/shadow rendering, base colors, projective distortions, natural image blending and noise. 

\textbf{SynthText (ST)} is a synthetically generated dataset made of 5.5M word images. SynthText blends synthetic text on natural images. It uses the scene geometry, texture, and surface normal to naturally blend and distort a text rendering on the surface of an object. The text is then cropped from the modified natural image.

\subsection{Test Dataset}

\begin{figure}
    \centering
    
\begin{center}
    
\setlength\tabcolsep{1pt}
\begin{tabular}{| r | c  c  |  r | c  c  |}

\hline
\multicolumn{3}{|c|}{\textbf{Regular Dataset} } & \multicolumn{3}{|c|}{\textbf{Irregular Dataset} } \\
\hline
IIIT5K 
&
\includegraphics[scale=0.18]{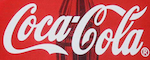} & 
\includegraphics[scale=0.18]{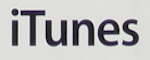} &
IC15
&
\includegraphics[scale=0.18]{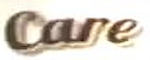}
&
\includegraphics[scale=0.18]{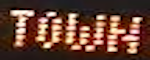}
\\ 

SVT 
&
\includegraphics[scale=0.18]{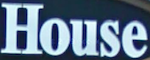} 
& 
\includegraphics[scale=0.18]{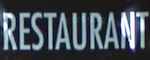} 
&
SVTP 
&
\includegraphics[scale=0.18]{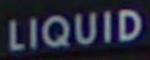} & 
\includegraphics[scale=0.18]{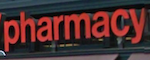} 
\\ 

IC03 
&
\includegraphics[scale=0.18]{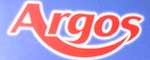} & 
\includegraphics[scale=0.18]{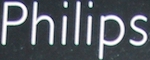} &

CT 
&
\includegraphics[scale=0.18]{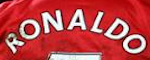} 
& \includegraphics[scale=0.18]{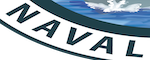} 
\\ 

IC13 
&
\includegraphics[scale=0.18]{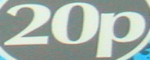} & 
\includegraphics[scale=0.18]{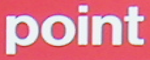} &

{}
& 
{}
&
{}

\\ 

\hline

\end{tabular}
\end{center}
    \caption{Sample text images from test datasets.}
    \label{fig:real_dataset}
\end{figure}

\begin{table}[]
    \centering
    \caption{Train conditions}
    \begin{tabular}{|l|l|}
    \hline
        \makecell{\textbf{Train dataset}: \\50\%MJ + 50\%ST} & \textbf{Batch size}: 192\\
    \hline
        \textbf{Iterations}: 300,000 & \textbf{Parameter init}:  He \cite{he2015delving} \\
    \hline
         \textbf{Optimizer}: Adadelta \cite{zeiler2012adadelta} & \textbf{Learning rate}: 1.0\\
    \hline
         \textbf{Adadelta $\rho$}: 0.95 & \textbf{Adadelta $\epsilon$}: $1e^{-8}$\\
    \hline
        \textbf{Loss}: Cross-Entropy/CTC & \textbf{Gradient clipping}: 5.0 \\
    \hline
        \textbf{Image size}: $100\times{32}$ & \textbf{Channels}: 1 (grayscale) \\
    \hline
    \end{tabular}
    \label{tab:train_condition}
\end{table}

\begin{table*}[]
    \centering
    \caption{Individual group absolute percent gain in accuracy for the RARE \cite{shi2016robust} model.}
    \small
    \begin{tabular}{| l |c | c | c | c | c | c | c | c | c | c | c |  }
    \hline
    \multirow{2}{*}{Group}  & IIIT & SVT & \multicolumn{2}{c|}{IC03} & \multicolumn{2}{c|}{IC13} & \multicolumn{2}{c|}{IC15} & SVTP & CT & Acc  \\
    {} & 3,000 & 647 & 860 & 867 & 857 & 1,015 & 1,811 & 2,077 & 645 & 288 & \%  \\
    \hline
    
    \textit{Warp} & \cellcolor{green}0.34	&\cellcolor{green}0.03	&\cellcolor{green}0.25	&\cellcolor{green}0.21	&\cellcolor{cyan}-0.12	&\cellcolor{green}0.08	&\cellcolor{green}0.78	&\cellcolor{green}0.51	&\cellcolor{cyan}-0.13	&\cellcolor{green}0.52	&\cellcolor{green}0.33 \\
    \hline
    \textit{Blur} & \cellcolor{cyan}-0.08	&\cellcolor{green}0.21	&\cellcolor{green}0.54	&\cellcolor{green}0.37	&\cellcolor{cyan}-0.10	&\cellcolor{cyan}-0.10	&\cellcolor{green}2.05	&\cellcolor{green}1.76	&\cellcolor{green}0.18	&\cellcolor{green}0.29	& \cellcolor{green}0.67 \\
    \hline
    \textit{Noise} & \cellcolor{cyan}-0.31	&\cellcolor{green}0.53	&\cellcolor{green}0.56	&\cellcolor{green}0.63	&\cellcolor{green}0.12	&0.00	&\cellcolor{green}1.49	&\cellcolor{green}1.29	&\cellcolor{green}1.10	&\cellcolor{cyan}-1.38	&\cellcolor{green}0.52 \\
    \hline
    \textit{Geometry} & \cellcolor{green}0.17	&\cellcolor{cyan}-0.18	&\cellcolor{green}0.10	&\cellcolor{green}0.13	&\cellcolor{green}0.39	&\cellcolor{green}0.33	&\cellcolor{green}0.99	&\cellcolor{green}0.87	&\cellcolor{cyan}-0.13	&\cellcolor{green}0.41	&\cellcolor{green}0.41 \\
    \hline
    \textit{Camera} & \cellcolor{cyan}-0.09	&\cellcolor{cyan}-0.18	&\cellcolor{green}0.03	&\cellcolor{green}0.03	&\cellcolor{cyan}-0.23	&\cellcolor{cyan}-0.06	&\cellcolor{green}1.16	&\cellcolor{green}0.93	&\cellcolor{green}0.26	&\cellcolor{cyan}-2.56	&\cellcolor{green}0.24\\
    \hline
    
    \textit{Weather} & \cellcolor{green}0.26	&\cellcolor{green}0.41	&\cellcolor{green}0.10	&\cellcolor{green}0.33	&\cellcolor{cyan}-0.54	&\cellcolor{cyan}-0.47	&\cellcolor{green}1.03	&\cellcolor{green}0.80	&\cellcolor{green}0.98	&\cellcolor{cyan}-0.13	&\cellcolor{green}0.38\\
    \hline
    \textit{Pattern} & \cellcolor{cyan}-0.35	&\cellcolor{cyan}-0.95	&0.00	&\cellcolor{cyan}-0.07	&\cellcolor{cyan}-0.33	&\cellcolor{cyan}-0.53	&\cellcolor{green}0.87	&\cellcolor{green}0.72	&\cellcolor{cyan}-0.45	&\cellcolor{cyan}-0.34	&\cellcolor{green}0.01\\
    \hline
    \textit{Process} & \cellcolor{cyan}-0.48	&\cellcolor{cyan}-0.57	&\cellcolor{green}0.27	&\cellcolor{green}0.18	&\cellcolor{cyan}-0.12	&\cellcolor{cyan}-0.07	&\cellcolor{green}1.05	&\cellcolor{green}0.76	&\cellcolor{green}0.31	&\cellcolor{green}0.64	&\cellcolor{green}0.19\\
    
    \hline
    \end{tabular}

    \label{tab:rare_ablation}
\end{table*}

The test dataset is made of several small publicly available STR datasets of text in natural images. These datasets are generally grouped into two: 1) Regular and 2) Irregular. 

The regular datasets have text images that are mostly frontal with a minimal amount of rotation or perspective distortion. IIIT5K-Words (IIIT) \cite{mishra2012scene}, Street View Text (SVT) \cite{wang2011end}, ICDAR2003 (IC03) \cite{lucas2005icdar} and ICDAR2013 (IC13) \cite{karatzas2013icdar} are considered regular datasets. \textbf{IIIT5K} has 3,000 test images. These images are mostly from street scenes such as sign boards, brand logos, house number or street signs. \textbf{SVT} has 647 test images. The text images are cropped from Google Street View images. \textbf{IC03} has 1,110 test images from ICDAR2003 Robust Reading Competition. Images were captured from natural scenes. Both versions, 860 and 867 test images, are used. \textbf{IC13} is an extension of IC03 and shares similar images. IC13 was created for the ICDAR2013 Robust Reading Competition.  Both versions, 857 and 1,015 test images, are used.

Meanwhile, irregular datasets are made of text with challenging appearances such as curved, vertical, under perspective transformation, low-resolution or distorted. ICDAR2015 (IC15) \cite{karatzas2015icdar}, SVT Perspective (SVTP) \cite{phan2013recognizing} and CUTE80 (CT) \cite{risnumawan2014robust} belong to irregular datasets. \textbf{IC15} has text images from the ICDAR2015 Robust Reading Competition. Many images are blurry, noisy, rotated, and sometimes of low-resolution, perspective-shifted, vertical and curved. Both versions, 1,811 and 2,077 test images, are used. \textbf{SVTP} has 645 test images from Google Street View. Most are images of business signage. \textbf{CT} focuses on curved text images captured from shirts and product logos. The dataset has 288 test images.

Figure \ref{fig:real_dataset} shows samples from both regular and irregular datasets. For both datasets, only the test splits are used in evaluating STR models.

\subsection{Experimental Setup}

The training configurations used in the framework are summarized in Table \ref{tab:train_condition}. We reproduced the results of 6 strong baseline models: CRNN \cite{shi2016end}, R2AM \cite{lee2016recursive}, GCRNN \cite{wang2017gated}, Rosetta \cite{borisyuk2018rosetta}, RARE \cite{shi2016robust} and TRBA \cite{baek2019wrong}. Each model is differentiated by 4 stages \cite{baek2019wrong}: 1) \textit{Image Rectification}: TPS \cite{jaderberg2015spatial} or None, 2) \textit{Feature Extractor}: VGG \cite{simonyan2014very}, ResNet \cite{he2016deep} or RCNN \cite{lee2016recursive}, 3) \textit{Sequence Modelling}: BiLSTM \cite{shi2016end} or None, and 4) \textit{Prediction}: Attention \cite{shi2016robust,cheng2017focusing} or CTC \cite{graves2006connectionist}. We trained all models from scratch for at least 5 times using different random seeds. The best performing weights on the test datasets are saved to get the mean evaluation scores. %Please note that Baek et al. \cite{baek2019wrong} mentioned that using the built-in \texttt{CTCLoss()}  in the recent versions of PyTorch instead of Baidu's \texttt{CTCLoss()} reduces the accuracy of CRNN, GRCNN and Rosetta by about 1\%. 

\subsection{Individual Group Performance}
After establishing the baseline scores, each STRAug group was used as a data augmentation method during training in order to understand the individual gain in accuracy. We performed an ablation study using the RARE model since it is the smallest model with all 4 stages present. Data augmentation is randomly applied with 50\% probability. The magnitude of data augmentation is randomly drawn from $(0, 1, 2)$. Table \ref{tab:rare_ablation} shows that the biggest gain of 0.67\% in absolute accuracy is from \textit{Blur}, followed by \textit{Noise} 0.52\%, \textit{Geometry} 0.41\%, \textit{Weather} 0.38\%, \textit{Warp} 0.33\%, \textit{Camera} 0.24\%, \textit{Process} 0.19\% and \textit{Pattern} 0.01\%. \textit{Blur} has the biggest gain on IC15 since the dataset has a substantial number of low resolution, low-light and blurry images. As expected, \textit{Warp} improves curved text that can be found in CT. Both \textit{Warp} and \textit{Geometry} improved the model performance on IC15 and IC03. Both datasets have rotated and distorted text. Surprisingly, SVTP did not improve with \textit{Warp} and \textit{Geometry}. We believe that while SVTP is from Google Street View, the amount of perspective distortion and rotation is not that significant unlike in IC15 and IC13. \textit{Noise} has performance gains across all datasets except for CT and IIIT. These two datasets are dominated by clean text images. \textit{Weather} has accuracy gains on both IC15 and SVTP. Both datasets have a substantial number of outdoor scenes. \textit{Process} improved the model performance on IC15, SVTP, CT and IC13. These datasets are characterized by highly varied color and texture. Overall, \textit{Pattern} has a negligible positive impact but it has a significant contribution in IC15 which contains text images with patterns. An example is the word TOWN in Figure \ref{fig:real_dataset}.

\begin{figure}
    \centering
    \includegraphics[scale=0.51]{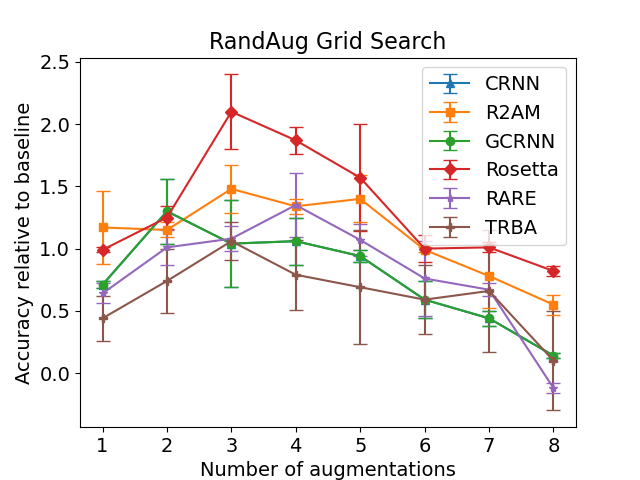} 
    \caption{Overall absolute percentage accuracy increase versus the number of data augmentations when STRAug is used in different baseline STR models.}
    \label{fig:randaug_grid}
\end{figure}

\begin{table*}[]
    \centering
    \caption{Reproduced baseline scores and absolute accuracy increase using different data augmentation methods.}
    \small
    \begin{tabular}{| l |r | r | r | r | r | r | r | r | r | r | r |  }
    \hline
    Model  &  \multicolumn{1}{c|}{IIIT}  & \multicolumn{1}{c|}{SVT} & \multicolumn{2}{c|}{IC03} & \multicolumn{2}{c|}{IC13} & \multicolumn{2}{c|}{IC15} & \multicolumn{1}{c|}{SVTP} & \multicolumn{1}{c|}{CT} & \multicolumn{1}{c|}{Acc}  \\
    
    +Augmentation & \multicolumn{1}{c|}{3,000} & \multicolumn{1}{c|}{647} & \multicolumn{1}{c|}{860} & \multicolumn{1}{c|}{867} & \multicolumn{1}{c|}{857} & \multicolumn{1}{c|}{1,015} & \multicolumn{1}{c|}{1,811} & \multicolumn{1}{c|}{2,077} & \multicolumn{1}{c|}{645} & \multicolumn{1}{c|}{288} & \multicolumn{1}{c|}{\%}  \\
    \hline
    
    CRNN \cite{shi2016end}&  
81.56	&80.26	&91.69	&91.41	&89.35	&88.30	&65.64	&60.80	&65.85	&61.63	&76.76    
    \\
    +SRN Aug &
\textbf{0.71}	&-0.23	&0.26	&0.26	&0.03	&0.02	&2.18	&1.81	&1.40	&2.69	&0.98    
\\
    +PP-OCR Aug &
0.37	&-0.14	&0.29	&0.25	&-0.01	&0.04	&2.11	&1.83	&0.56	&\textbf{3.99}	&0.88    
\\
    +STRAug (Ours) &
0.63	&\textbf{0.37}	&\textbf{0.79}	&\textbf{0.87}	&\textbf{0.22}	&\textbf{0.24}	&\textbf{2.68}	&\textbf{2.34}	&\textbf{2.00}	&2.26	&\textbf{1.30}
    \\
    \hline
    R2AM \cite{lee2016recursive}&
83.28	&80.95	&91.69	&91.38	&90.17	&88.15	&68.48	&63.35	&70.50	&64.67	&78.46
\\
    +SRN Aug &
0.81	&2.72	&\textbf{0.99}	&\textbf{1.01}	&0.92	&0.68	&1.46	&1.49	&1.80	&2.11	&1.24
\\
    +PP-OCR Aug &
\textbf{0.84}	&1.17	&\textbf{0.99}	&0.82	&\textbf{1.04}	&\textbf{1.04}	&1.90	&1.78	&1.49	&\textbf{4.08}	&1.33
\\
    +STRAug (Ours) &
\textbf{0.84}	&\textbf{2.92}	&0.60	&0.59	&0.73	&0.62	&\textbf{2.34}	&\textbf{2.10}	&\textbf{2.67}	&3.15	&\textbf{1.48}
\\
    \hline
    GCRNN \cite{wang2017gated}&
82.89	&\textbf{81.14}	&\textbf{92.67}	&92.31	&89.97	&88.37	&68.12	&62.94	&68.48	&65.51	&78.30
\\
    +SRN Aug &
\textbf{0.54}	&\textbf{0.00}	&-0.35	&-0.42	&-0.08	&0.53	&0.35	&0.42	&0.21	&2.20	&0.31
\\
+PP-OCR Aug &
0.34	&\textbf{0.00}	&-0.23	&-0.15	&-0.39	&0.16	&0.96	&0.91	&1.40	&\textbf{3.13}	&0.49
\\
+STRAug (Ours) &
0.52	&-0.41	&-0.12	&\textbf{0.04}	&\textbf{0.31}	&\textbf{0.79}	&\textbf{1.73}	&\textbf{1.62}	&\textbf{2.22}	&1.62	&\textbf{0.89}
\\
\hline
    Rosetta \cite{borisyuk2018rosetta}&
82.59	&82.60	&92.60	&91.97	&90.32	&88.79	&68.15	&62.95	&70.02	&65.76	&78.43
\\
    +SRN Aug &
1.06	&0.14	&0.46	&0.49	&-0.12	&0.05	&2.58	&2.44	&1.10	&\textbf{5.76}	&1.34    
\\
+PP-OCR Aug &
2.14	&0.97	&\textbf{0.69}	&\textbf{0.76}	&0.39	&0.28	&2.58	&2.44	&1.46	&4.72	&1.74
\\
+STRAug (Ours) &
\textbf{2.15}	&\textbf{1.58}	&0.47	&0.72	&\textbf{0.61}	&\textbf{0.69}	&\textbf{3.47}	&\textbf{3.30}	&\textbf{1.61}	&5.07	&\textbf{2.10}
\\
\hline
    RARE \cite{shi2016robust}&
85.95	&85.19	&93.51	&93.33	&92.30	&91.03	&73.94	&68.42	&75.58	&70.54	&82.12
\\
+SRN Aug &
0.18	&0.34	&\textbf{0.91}	&\textbf{0.87}	&\textbf{0.62}	&0.59	&2.04	&1.74	&0.96	&1.22	&0.97
\\
+PP-OCR Aug &
0.16	&0.95	&0.10	&-0.13	&0.51	&0.62	&\textbf{2.43}	&\textbf{2.16}	&\textbf{2.14}	&2.49	&1.09
\\
+STRAug (Ours) &
\textbf{0.75}	&\textbf{1.31}	&0.83	&0.67	&\textbf{0.62}	&\textbf{0.72}	&\textbf{2.43}	&2.06	&1.63	&\textbf{2.95}	&\textbf{1.35}
\\
\hline
    TRBA \cite{baek2019wrong}&
87.71	&87.44	&94.54	&94.20	&93.38	&92.14	&77.32	&71.62	&78.14	&75.52	&84.29
\\
+SRN Aug & 
% For update
0.85	&\textbf{0.72}	&-0.02	&0.21	&0.16	&0.31	&2.00	&1.82	&1.74	&0.80	&1.02
\\
+PP-OCR Aug &
0.70	&0.23	&0.17	&0.00	&0.32	&0.54	&\textbf{2.03}	&\textbf{1.83}	&\textbf{2.02}	&2.08	&1.04
\\
+STRAug (Ours) &
% For update
\textbf{1.23}	&0.58	&\textbf{0.35}	&\textbf{0.52}	&\textbf{0.64}	&\textbf{0.69}	&1.35	&1.08	&1.94	&\textbf{2.60}	&\textbf{1.06}
\\
\hline
\hline
    ViTSTR-Tiny \cite{atienza2021vision}&
83.7	&83.2	&92.8	&92.5	&90.8	&89.3	&72.0	&66.4	&74.5	&65.0	&80.3
\\
    ViTSTR-Tiny+STRAug&
85.1	&85.0	&93.4	&93.2	&90.9	&89.7	&74.7	&68.9	&78.3	&74.2	&82.1
\\
ViTSTR-Small &
85.6	&85.3	&93.9	&93.6	&91.7	&90.6	&75.3	&69.5	&78.1	&71.3	&82.6
\\
ViTSTR-Small+STRAug&
86.6	&87.3	&94.2	&94.2	&92.1	&91.2	&77.9	&71.7	&81.4	&77.9	&84.2	\\
   ViTSTR-Base
 &86.9	&87.2	&93.8	&93.4	&92.1  &91.3	&76.8	&71.1	&80.0	&74.7	&83.7
\\
ViTSTR-Base+STRAug
&88.4	&87.7	&94.7	&94.3	&93.2	&92.4	&78.5	&72.6	&81.8	&81.3	&85.2
\\
\hline
    \end{tabular}

    \label{tab:comparison}
\end{table*}

\subsection{Combined Group Performance}
The individual group performance gains may not appear impressive. However, combining all groups significantly pushes the accuracy higher. We used RandAugment \cite{cubuk2020randaugment} as a policy to randomly select $N$ data augmentation groups each with a random magnitude $M$ to apply during training. Unlike AutoAugment \cite{cubuk2019autoaugment}, RandAugment has a comparable performance and is easy to optimize using a simple grid search. Figure \ref{fig:randaug_grid} shows the absolute percent accuracy gain versus $N$ for different baseline models. Unlike in object recognition where the accuracy increases with the number of augmentations, in STR the peak is between $N=2$ and $N=4$. Table \ref{tab:comparison} shows the gains per dataset. When compared with the results from applying data augmentation on an individual group basis, the gains using a mixture of groups are significantly higher. For example, in evaluating CT with the RARE model the highest group gain is only 0.64\% while it is 2.95\% or about 4.6$\times$ for the combined.

Given the library of STRAug functions, it is easy to implement and validate other data augmentation algorithms. For example, SRN \cite{yu2020towards} data augmentation which is made of random resizing plus padding, rotation, perspective transformation, motion blur and Gaussian noise can be formulated as:

\footnotesize
\noindent
\texttt{geometry = [Rotate(), Perspective(), Shrink()]} 
%\indent\indent\indent\indent\indent  

\noindent
\texttt{noise = [GaussianNoise()]}

\noindent
\texttt{blur = [MotionBlur()]}

\noindent
\texttt{augmentations = [geometry, noise, blur]}

\noindent
\texttt{img = RandAugment(img, augmentations, N=3)}

\normalsize
Similarly, using STRAug functions, we can easily implement and validate PP-OCR \cite{du2020pp} data augmentation. PP-OCR uses the combined methods of SRN and \textit{Learn to Augment} (i.e. random distortion). Note that the main difference of our implementation of SRN and PP-OCR data augmentations is that we further fine tuned both methods using RandAugment to maximize their potential.

Table \ref{tab:comparison} presents a comparison of the absolute accuracy increase due to SRN, PP-OCR and STRAug data augmentation techniques in the baseline models. The increase is significant especially on challenging irregular datasets such as CT (1.62\%-5.07\%), SVTP (1.61\%-2.67\%) and IC15 (1.08\%-3.47\%). Figure \ref{fig:straug_teaser} shows example text images that baseline models made correct predictions when trained with STRAug. Given that STRAug is using a more diverse set of data augmentation functions, it outperforms recent STR data augmentation methods. STRAug is also an effective regularizer on a vision tranformer-based STR such ViTSTR\cite{atienza2021vision}. Table  \ref{tab:comparison} shows substantial gains in performance on all sizes, Tiny: +1.8\%, Small: +1.6\% and Base: +1.5\%.

\section{Conclusion}
STRAug is a library of diverse 36 STR data augmentation functions with a simple API. The empirical results showed that a significant accuracy gain can be obtained using STRAug. 

\section{Acknowledgement}
This work was funded by the University of the Philippines ECWRG 2019-2020. Thanks to CNL people: Roel Ocampo and Vladimir Zurbano, for hosting our servers. 

%-------------------------------------------------------------------------

%-------------------------------------------------------------------------

%------------------------------------------------------------------------

{\small
\bibliographystyle{ieee_fullname}
\bibliography{egpaper_final}
}

\end{document}